%% file: main.tex
\documentclass[letterpaper, 10 pt,journal,twoside]{IEEEtran}

\IEEEoverridecommandlockouts                              

\usepackage{graphics} 
\usepackage[colorlinks=true,allcolors=black]{hyperref}
\usepackage{stmaryrd} 
\usepackage{epsfig} 

\usepackage{times} 
\usepackage{amsmath} 
\usepackage{amssymb}  
\usepackage{float}
\usepackage{hyperref}
\usepackage{subcaption}
\usepackage{physics}
\usepackage{tikz}
\usepackage{pgfplots}
\usepackage{balance}
\usepackage{tabu}
\usepackage{multirow,booktabs}
\usepackage{cite}
\usepackage[linesnumbered,ruled,vlined]{algorithm2e}

\newcommand{\sref}[1]{{Section \ref{#1}}}
\newcommand{\fref}[1]{Fig. \ref{#1}}
\newcommand{\tref}[1]{Table \ref{#1}}
\newcommand{\aref}[1]{Algorithm \ref{#1}}
\title{
Lightweight 3-D Localization and Mapping for Solid-State LiDAR}

\author{Han Wang$^{1}$, Chen Wang$^{2}$, and Lihua Xie$^{1}$
\thanks{Manuscript received: October, 15, 2020; Revised December, 9, 2020; Accepted January, 30, 2021.}
\thanks{This paper was recommended for publication by Editor Sven Behnke upon evaluation of the Associate Editor and Reviewers' comments.
This work was supported by Delta-NTU Corporate Laboratory for Cyber-Physical Systems under the National Research Foundation Corporate Lab @ University Scheme.} 
\thanks{$^{1}$Han Wang and Lihua Xie are with the School of Electrical and Electronic Engineering,
Nanyang Technological University, 50 Nanyang Avenue, Singapore 639798.
        {\tt\small e-mail: \{wang.han,elhxie\}@ntu.edu.sg}}
\thanks{$^{2}$Chen Wang is with the Robotics Institute, Carnegie Mellon University, Pittsburgh, PA 15213, USA. {\tt\small e-mail: chenwang@dr.com}}
\thanks{Digital Object Identifier (DOI): see top of this page.}
}

\begin{document}
 
\maketitle

\begin{abstract}

\input{body/abstract.tex}

\end{abstract}
\section{INTRODUCTION}
\input{body/Introduction.tex}

\section{RELATED WORK}\label{sec:related-work}
\input{body/RelatedWork.tex}

\section{METHODOLOGY}\label{sec:methodology}
\input{body/Methodology.tex}

\section{EXPERIMENT EVALUATION}\label{sec:experiment}
\input{body/Experiments.tex}

\section{CONCLUSION}\label{sec:conclusion}
\input{body/Conclusion.tex}

\balance
\bibliographystyle{IEEEtran}
\bibliography{IEEEabrv,references}

\end{document}

%% file: body/abstract.tex
The LIght Detection And Ranging (LiDAR) sensor has become one of the most important perceptual devices due to its important role in simultaneous localization and mapping (SLAM). Existing SLAM methods are mainly developed for mechanical LiDAR sensors, which are often adopted by large scale robots. Recently, the solid-state LiDAR is introduced and becomes popular since it provides a cost-effective and lightweight solution for small scale robots.
Compared to mechanical LiDAR, solid-state LiDAR sensors have higher update frequency and angular resolution, but also have smaller field of view (FoV), which is very challenging for existing LiDAR SLAM algorithms.
Therefore, it is necessary to have a more robust and computationally efficient SLAM method for this new sensing device. To this end, we propose a new SLAM framework for solid-state LiDAR sensors, which involves feature extraction, odometry estimation, and probability map building.
The proposed method is evaluated on a warehouse robot and a hand-held device. In the experiments, we demonstrate both the accuracy and efficiency of our method using an Intel L515 solid-state LiDAR. The results show that our method is able to provide precise localization and high quality mapping. We made the source codes public at \url{https://github.com/wh200720041/SSL_SLAM}.


%% file: body/Introduction.tex
\IEEEPARstart{T}{he}  LIght Detection And Ranging (LiDAR) is a kind of Time of Flight (ToF) sensor which measures the object distance by emitting laser to the object and capturing the travelling time. It is one of the most popular perceptual systems in robotic applications due to its high precision, long cover range, and high reliability. 
Traditional localization approaches often leverage on external setup and hence lack of robustness in different scenarios.
For example, UWB localization \cite{nguyen2019persistently} relies on the pre-installation and calibration of multiple anchors, and WiFi localization \cite{zou2017winips} requires multiple routers in order to achieve high accuracy. In comparison, LiDAR SLAM provides an external device/landmark-free localization for mobile robots in both indoor and outdoor environments \cite{hong2017probabilistic,ye2019tightly, deschaud2018imls}. Moreover, LiDAR SLAM is less affected by environment uncertainties, \textit{e.g.}, weather change or illumination change, compared to other SLAM system such as visual SLAM \cite{mur2015orb, nguyen2020loosely} and visual inertial SLAM \cite{leutenegger2015keyframe, qin2018vins}. Those properties have made LiDAR SLAM widely applied to various robotic applications such as autonomous driving \cite{pfrunder2017real}, building inspection \cite{bircher2015structural}, and intelligent manufacturing \cite{intensity2020wang}. 

Existing approaches were mainly developed for mechanical LiDAR sensors which collect the surrounding information by spinning a high frequency laser array. Although they have achieved impressive experimental results on large scale mapping \cite{liosam2020shan, zhang2015visual}, it is not popularly used due to the high cost. Moreover, the mechanical LiDAR is difficult to be implemented on small scale system due to its size and weight. For example, the durability of flight of Unmanned Aerial Vehicles (UAVs) is significantly reduced by carrying a mechanical LiDAR for building inspection \cite{wallace2012development}. In addition, it is also not possible to integrate mechanical LiDAR into hand-held device due to its large size. 

\begin{figure}[t]
\begin{center}
\vspace{8pt}
\includegraphics[width=0.99\linewidth]{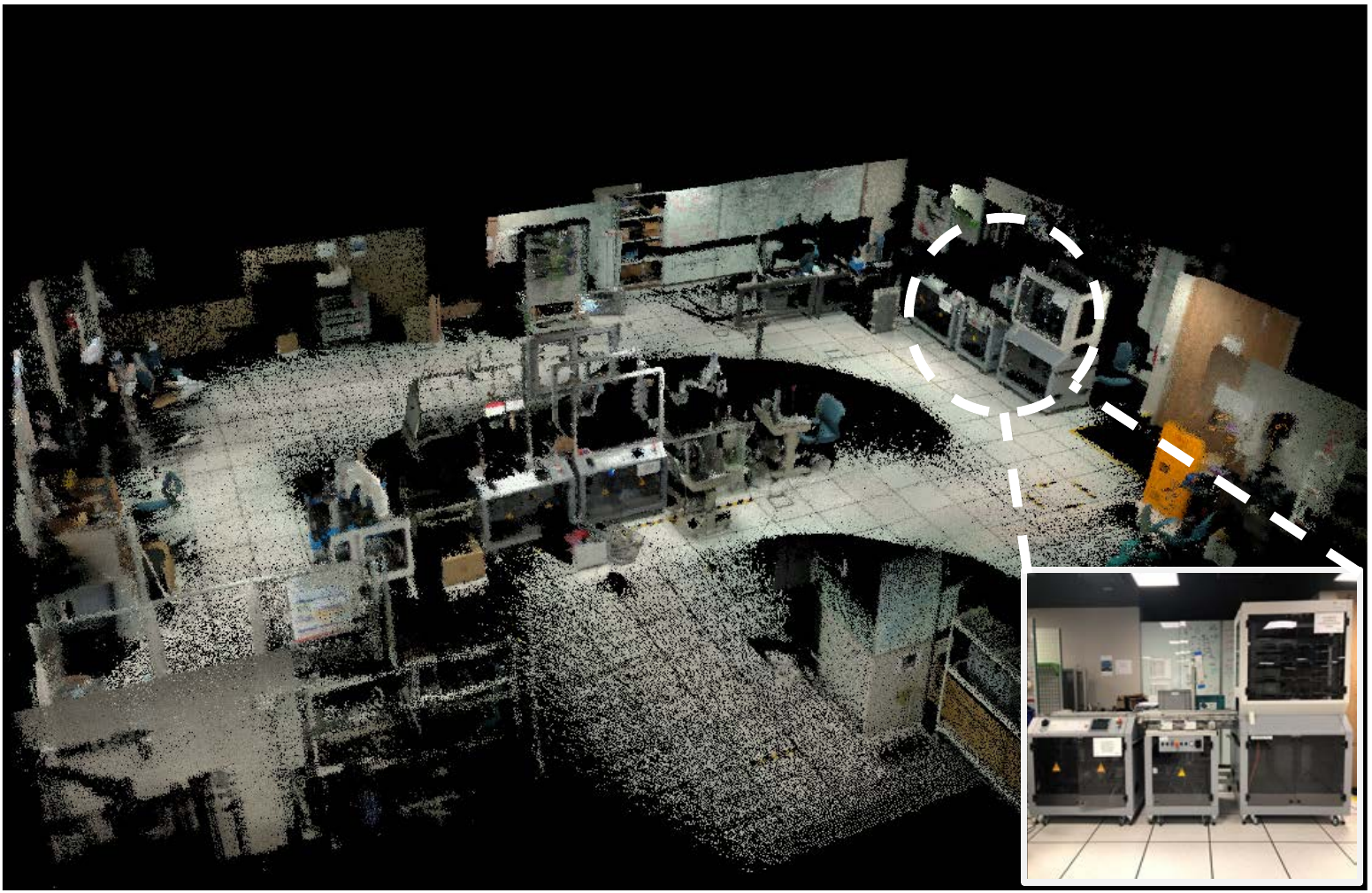}
\captionsetup{justification=justified}
\caption{Example of indoor localization and mapping in the warehouse environment. The proposed method is integrated to an AGV used for warehouse manipulation. Our method is able to provide real-time localization and dense mapping on an embedded mini computer.}
\label{fig: title_graph}
\end{center}
\end{figure}

\begin{table*}[t]
    \begin{center}
    \begin{tabular}{cccccccccc}
    \toprule
    
    \multirow{2}{*}{Sensor} & \multirow{2}{*}{Type} & \multirow{2}{*}{Frequency} &\multirow{2}{*}{FoV} & Horizontal & Vertical & Detection & \multirow{2}{*}{Accuracy} & \multirow{2}{*}{Dimension} & \multirow{2}{*}{Weight} \\
    &&&&Resolution&Resolution&Range&&\\
    \midrule
    Velodyne VLP-16 & Mechanical &5-20Hz& $360^{\circ} \times 32^{\circ}$    &  0.1-0.4$^{\circ}$ &$2.0^{\circ}$ & 0.5-100 m & 3 cm & 103 $\times$ 72 mm & 860 g\\  
    Realsense L515 & Solid-state & 30Hz &$70^{\circ} \times 55^{\circ}$ &  $0.07^{\circ}$ & $0.07^{\circ}$ &0.25-9 m & 1.4 cm & 61$\times$ 26 mm &95 g\\  
    \bottomrule
    \end{tabular}
    \caption{Difference of Mechanical and Solid-state LiDAR.}
    \label{table: Lidar comparison}
    \end{center}
\end{table*}

Recently, the introduction of solid-state LiDAR provides a cost-effective and lightweight solution for LiDAR SLAM system. The solid-state LiDAR is a system that built entirely on a silicon chip with no moving parts involved \cite{lin2020loam}. Hence both the size and the weight can be made much smaller than mechanical LiDAR. Moreover, the solid-state LiDAR is resistant to vibrations by removing rotating mechanical structure. The latest solid-state LiDAR only costs 10\% price of a mechanical LiDAR and is as small as a smart phone, which has a great potential to become a dominant perception system for small scale application such as Virtual Reality (VR) \cite{burdea2003virtual}, drone inspection, and indoor navigation. In terms of performance, solid-state LiDAR is able to achieve an accuracy up to 1-2 \textit{cm} and a detectable range up to a few hundreds of meters.

Although the performance of mechanical LiDAR and solid-state LiDAR is similar, the implementation or challenge of LiDAR SLAM is different. To illustrate the difference between two LiDARs, we use Velodyne VLP-16 and Realsense L515 for example. The specifications can be found at \tref{table: Lidar comparison}. Solid-state LiDAR has higher angular resolution, implying that the points are more dense within the same scanning area. Therefore, traditional LiDAR odometry methods such as Iterative Closest Point (ICP) \cite{besl1992method} can be computationally inefficient since there are more points involved for calculation. Secondly, the update frequency of solid-state LiDAR is higher, while traditional LiDAR SLAM such as LOAM \cite{zhang2014loam} is not computationally efficient enough to reach real-time performance. Another challenge is the pyramid-like coverage view that can cause severe tracking loss during large rotation. Therefore, a more computationally efficient and robust algorithm has to be designed in order to achieve real-time performance for solid-state LiDAR-based SLAM.

In this work, we propose a novel and lightweight SLAM framework for solid-state LiDAR, consisting of feature extraction, odometry estimation and probability map construction. Inspired by existing LiDAR SLAM methods such as LOAM \cite{zhang2014loam} and LeGO-LOAM \cite{shan2018lego}, we propose a new rotation-invariant feature extraction method that exploits both horizontal and vertical curvature. The proposed method provides real-time localization on mobile platforms. Thorough experiments have been conducted to evaluate its performance.
The main contributions of this paper are listed as follows:
\begin{itemize}
\item We propose a full SLAM framework for solid-state LiDAR, which targets to tackle perception system with small FoV and high updating frequency. We make the proposed method open sourced.
\item We introduce an improved feature extraction strategy which is able to search for consistent features under significant rotations. Moreover, left Lie derivative is used for iterative pose estimation so that the pose is stored in a singularity-free format. 
\item A thorough evaluation of the proposed method is presented. More specifically, we integrate an Intel L515 solid state LiDAR to AGVs and test the proposed method in a complex warehouse environment. The proposed method can provide real time localization and is robust under rotations.
\end{itemize}

This paper is organized as follows: \sref{sec:related-work} reviews the related works on existing LiDAR SLAM approaches. \sref{sec:methodology} describes the details of the proposed approach, including feature point selection, laser odometry estimation, and probability map construction. \sref{sec:experiment} shows experimental results and comparison with existing works, followed by conclusion in \sref{sec:conclusion}.

%% file: body/RelatedWork.tex
Existing works on LiDAR SLAM mainly estimate the location by either scan-to-scan matching or feature-based matching. To match between two scan inputs, Iterative Closest Point (ICP) \cite{chetverikov2002trimmed} is one of the most classic methods used. Starting with an initial pose guess, the ICP algorithm determines the point correspondence by finding the closest point in target frames. A transformation matrix between the current frame and target frame can be estimated by minimizing the Euclidean distance between point pairs. Subsequently, the derived transformation matrix leads to new correspondence relationship and new transformation matrix. And the transformation matrix converges through an iterative manner. A LiDAR scan input consists of tens of thousands of points and direct point cloud matching is computationally inefficient. Hence it is mainly implemented to LiDAR SLAM with high computational resources and low update frequency, \textit{e.g.}, HDL-Graph-SLAM \cite{koide2019portable}, and LiDAR-Only Odometry and Localization (LOL) \cite{rozenberszki2020lol}. In the meantime, raw point cloud matching is also sensitive to noise in practice. For example, in autonomous driving, the measurements from trees alongside the road are often noisy. Such noise greatly reduces the matching accuracy and causes localization drift.

\begin{figure*}[t]
\begin{center}
\vspace{5pt}
\includegraphics[width=0.69\linewidth]{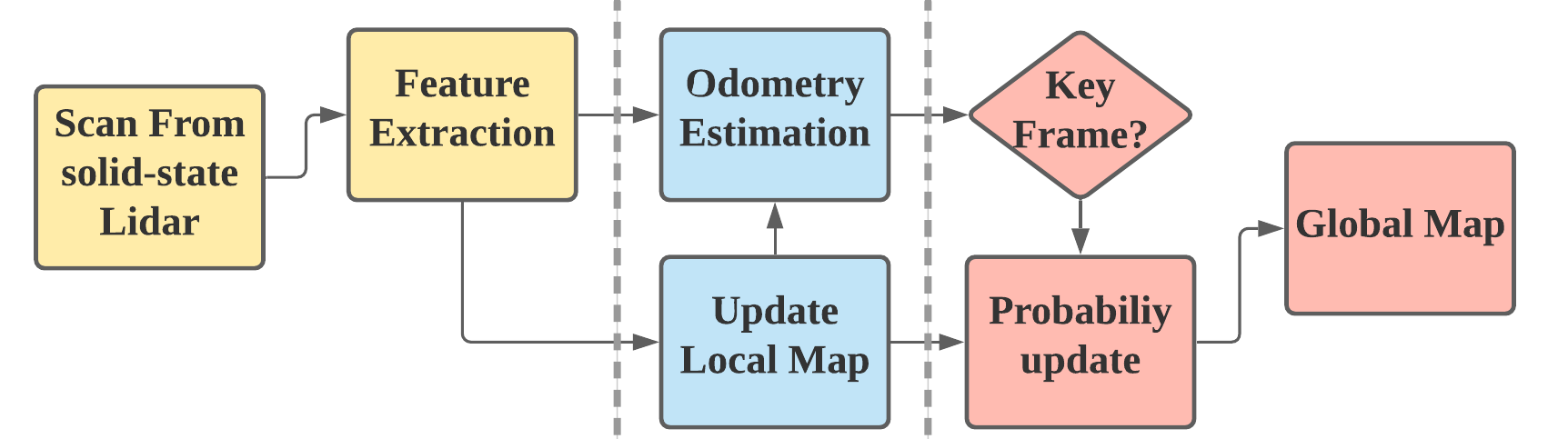}
\captionsetup{justification=justified}
\caption{System overview of the proposed method. It consists of three main modules: feature extraction, odometry estimation and probability map construction, which are highlighted in yellow, blue and red, respectively.}
\label{fig: flowchart}
\end{center}
\end{figure*}

A more computationally efficient way is to match in feature space where less points are involved to find the pose transformation. Zhang \textit{et al.} \cite{zhang2014loam} propose a new framework via matching on the edge features and planar features to solve the LiDAR SLAM problem for mobile robots. The edge features and planar features are collected based on local smoothness and are maintained in separate maps. And the feature correspondence is picked by finding the nearest global edges and planes respectively. The localization is estimated by minimizing the point-to-edge and point-to-plane distance. The experimental results also achieve impressive performance on public dataset evaluation such as KITTI dataset \cite{geiger2013vision}. An improved version is LeGO-LOAM \cite{shan2018lego}, which targets to solve the LiDAR SLAM for Unmanned Ground Vehicles (UGVs). The surrounding points from the current frame are segmented into ground points and non-ground points. For UGVs, the ground points are often planar points and those points are used to identify roll, pitch angle and z translation. The results are subsequently used as prior knowledge for edge feature matching, which targets to calculate x, y translation and yaw angle. By taking segmentation and two-stage matching, the feature points selected are reduced by 40\%. The experimental results show that LeGO-LOAM is able to run on small computing platform at real-time performance. 

Recently some works aim to use some specially designed LiDAR with small FoV. In \cite{lin2020loam} the author proposes a new LiDAR SLAM framework for small Fov LiDAR, \textit{i.e.}, the DJI Livox. DJI Livox spins a laser beam in cone shape direction and provides smaller FoV but denser points. To tackle the localization problem for small FoV sensor, the author introduces a new feature extraction method with the analysis of local LiDAR intensity. 
Compared to traditional feature extraction, the proposed feature extraction is more consistent over nearby frames which can subsequently reduce the tracking loss. However, it is mainly designed for large scale outdoor environment, while the performance in the indoor complex environment is not demonstrated.

%% file: body/Methodology.tex

In this section, the proposed method is described in details. As shown \fref{fig: flowchart}, the system consists of three main modules, namely feature extraction, odometry estimation, and probability map construction. We firstly present the rotation invariant feature extraction method and then introduce odometry estimation via local feature matching. Finally we present the probability map building and scene reconstruction.

\subsection{Feature Extraction}

A solid-state LiDAR integrates all sensors on a single silicon chip without moving parts. It often has higher resolution and higher update frequency compared to a mechanical LiDAR. Hence it might be too computational heavy to match the raw point clouds. Inspired by LOAM \cite{zhang2014loam}, we leverage the edge and planar matching which is much more computationally efficient. Before processing the data, we remove the noisy points based on the measured distance. It is observed that readings near the maximum detection range are often less accurate due to the low reflected energy thus we pre-filter those noisy points.


The point cloud from a LiDAR scan is unordered. To compute the edge and planar features, We firstly project the point cloud into a 2-D point matrix. 
For the $k^{th}$ LiDAR scan input $\mathcal{P}_{k}$, it is segmented based on the vertical angle and horizontal angle of each point. Given a point $\mathbf{p}_i = \{x_i,y_i,z_i\} \in \mathcal{P}_{k}$, the vertical angle $\alpha_i$ and horizontal angle $\theta_i$ are calculated by: 
\begin{equation}
\begin{aligned}
   \alpha_i &= \arctan \left(\frac{y_i}{x_i}\right),\\
    \theta_i &= \arctan \left(\frac{z_i}{x_i}\right).\\
\end{aligned}
\end{equation}
The point cloud is then segmented by equally diving vertical detection range $[\alpha_{min}, \alpha_{max}]$ and horizontal detection range $[\theta_{min}, \theta_{max}]$ into $M$ sectors and $N$ sectors, where $[\alpha_{min}$, $\alpha_{max}$, $\theta_{min}$ and $\theta_{max}$ are the minimum vertical angle, maximum vertical angle, minimum horizontal angle, maximum horizontal angle according to the sensor 
specifications. Hence, the point cloud is segmented into $M \times N$ cells. For a solid-state LiDAR with vertical angular resolution of $\alpha_r$ and horizontal resolution of $\theta_r$, $M$ and $N$ is selected as half of total points in a single direction, i.e., $M = \frac{\alpha_{max}-\alpha_{min}}{2 \times \alpha_r}$ and $N = \frac{\theta_{max}-\theta_{min}}{2 \times \theta_r}$. In general, larger $M$ and $N$ will take more computational resources. In each cell $(m,n)$, where $m \in [1,M]$, $n \in [1,N]$ and the symbol $[1,N]$ represents $\{1,2,\cdots,N\}$, we calculate the mean measurement $\textbf{p}_k^{(m,n)}$ by finding the geometry center of all points in the cell. 

To extract the edge and planar features, we search for its nearby points and define the local smoothness by:
\begin{equation}
  \sigma_{k}^{(m,n)} = \frac{1}{ \lambda^2}\cdot \sum_{\textbf{p}_{k}^{(i,j)} \in \mathcal{S}_{k}^{(m,n)}}(||\textbf{p}_{k}^{(i,j)}|| - ||\textbf{p}_{k}^{(m,n)}||),
\end{equation}
with
\begin{equation}
  \mathcal{S}_{k}^{(m,n)} = \{\textbf{p}_{k}^{(i,j)}| i\in [m-\lambda,m+\lambda],j\in [n-\lambda,n+\lambda]\},
\end{equation}
where $\lambda$ is a pre-defined searching radius. 
A larger $\lambda$ means searching for more surrounding points and requires more computational resource. 
The local smoothness indicates the sharpness of surrounding information.
A higher $\sigma_{k}^{(m,n)}$ indicates that the local surface is sharp, thus the corresponding point $\textbf{p}_{k}^{(m,n)}$ is taken as edge feature. 
A smaller $\sigma_{k}^{(m,n)}$ indicates that the local surface is flat, thus the corresponding point $\textbf{p}_{k}^{(m,n)}$ is taken as plane feature. 

\subsection{Odometry Estimation}
The odometry estimation is a task to estimate the robot current pose $\textbf{T}_{k} \in SE(3)$ in global coordinate based on the historical laser scan $\mathcal{P}_1,\mathcal{P}_2, \cdots, \mathcal{P}_{k-1}$. Traditionally the trajectory is estimated by either scan-to-scan match or scan-to-map match. The scan-to-scan match aligns the current frame with last frame that is faster. However, a single laser scan contains less surrounding information compared to local map and causes drift in the long run. Therefore, we use the scan-to-map match in order to improve the performance. To reduce the computational cost, sliding window approach is used. The edge features and planar features from the nearby frames are used to build the local feature maps. For the current input $P_{k}$, we define the local map $M_{k} = \{P_{k-1},P_{k-2},\cdots,P_{k-q}\}$, where $q$ is the number of frames used to build local map. 

As mentioned above, matching on raw point clouds is less efficient and sensitive to noise. Thus, we leverage in matching edge points and planar points in feature space. For each edge point $\textbf{p}_k\in P_k$ and its transform in the local map coordinate $\hat{\textbf{p}}_k = \textbf{T}_{k}\cdot \textbf{p}_k$, we search for the nearest edge from the local map. Note that the local map is divided into edge map and planar map, and each map is built by K-D tree in order to increase searching efficiency. Thus we can select two nearest edge feature points $\textbf{p}^{\mathcal{E}}_{1}$ and $\textbf{p}^{\mathcal{E}}_{2}$ from the local map for each edge point. The point-to-edge residual $f_{\mathcal{E}}(\hat{\textbf{p}}_{k})$ is defined as the distance between $\hat{\textbf{p}}_{k}$ and the edge crossing $\textbf{p}^{\mathcal{E}}_{1}$ and $\textbf{p}^{\mathcal{E}}_{2}$: 
\begin{equation}
  f_{\mathcal{E}}(\hat{\textbf{p}}_{k}) = \frac{|(\hat{\textbf{p}}_{k} - \textbf{p}^{\mathcal{E}}_2)\times (\hat{\textbf{p}}_{k} - \textbf{p}^{\mathcal{E}}_1)|}{|\textbf{p}^{\mathcal{E}}_1-\textbf{p}^{\mathcal{E}}_2|},
\end{equation}
where symbol $\times$ is the cross product of two vectors. Note that if the number of nearby points is less than 2, the point-to-edge residual is not taken into accounted by the final cost. Similarly, for each planar point $\textbf{p}_k\in P_k$, we search for the nearest planar features from the local map. To estimate a plane in 3D space, it is necessary to have 3 points. Hence for a given planar feature point $\textbf{p}_{k}$ and its transform in local map coordinate $\hat{\textbf{p}}_k = \textbf{T}_{k}\cdot \textbf{p}_k$, we can find 3 nearest points $\textbf{p}_{\mathcal{S}}^{1}$, $\textbf{p}_{\mathcal{S}}^{2}$, and $\textbf{p}_{\mathcal{S}}^{3}$ from the local map. The point-to-plane residual $f_{\mathcal{S}}(\hat{\textbf{p}}_{k})$ is defined as the distance between $\hat{\textbf{p}}_{k}$ and the plane crossing $\textbf{p}_{\mathcal{S}}^{1}$, $\textbf{p}_{\mathcal{S}}^{2}$, and $\textbf{p}_{\mathcal{S}}^{3}$: 
\begin{equation}
  f_{\mathcal{S}}(\hat{\textbf{p}}_k) = |(\hat{\textbf{p}}_{k} - \textbf{p}^{\mathcal{S}}_1)^T \cdot \frac{(\textbf{p}^{\mathcal{S}}_1 - \textbf{p}^{\mathcal{S}}_2)\times (\textbf{p}^{\mathcal{S}}_1-\textbf{p}^{\mathcal{S}}_3)}{|(\textbf{p}^{\mathcal{S}}_1 - \textbf{p}^{\mathcal{S}}_2)\times (\textbf{p}^{\mathcal{S}}_1-\textbf{p}^{\mathcal{S}}_3)|}|.
\end{equation}
Similar to the edge residual, the point-to-plane residual is not taken into account when the number of nearby points is less than 3. The final odometry is estimated by minimizing the point-to-plane residual and point-to-edge residual:
\begin{equation}
   \text{arg}\min_{\textbf{T}_{k}}  \sum_{} f_{\mathcal{E}}(\hat{\textbf{p}}_k) + \sum_{}f_{\mathcal{S}}(\hat{\textbf{p}}_k).
\end{equation}
This non-linear optimization problem can be solved by the Gauss–Newton optimization method. We use left perturbation scheme and apply increment on Lie Group. Compared to the differentiation model in LOAM \cite{zhang2014loam}, there are several advantages: (i) the rotation or pose is stored in a singularity-free format; (ii) at each iteration unconstrained optimization is performed; (iii) the manipulations occur at the matrix level so that there is no need to worry about taking the derivatives of a bunch of scalar trigonometric functions, which can easily lead to errors \cite{barfoot2016state}. Define $\xi_{k}=[\rho,\phi] \in \mathfrak{se}(3)$ and the transformation matrix:
\begin{equation}
   \textbf{T}_k = \exp(\xi_k^{\wedge}),
\end{equation}
where the notation $\xi^{\wedge}$ converts a 6D vector into a $4\times 4$ matrix by:
\begin{equation}
\begin{aligned}
  \xi^{\wedge} = \begin{bmatrix}
  [\rho]_{\times} &\phi\\
  \textbf{0}_{1\times 3} &0
  \end{bmatrix},
\end{aligned}
\end{equation}
where $[\cdot]_{\times}$ is the skew matrix of a 3D vector. The left perturbation model can be calculated by:
\begin{equation}
\begin{aligned}
 \textbf{J}_p = \frac{\partial \textbf{T}_k\textbf{p}_k}{\partial \delta\xi} & =\lim_{\delta \xi \to \textbf{0}} \frac{(\exp({\delta \xi^\wedge)}\cdot \textbf{T}_k\textbf{p}_k - \textbf{T}_k\textbf{p}_k)}{\delta \xi} \\
    &= \begin{bmatrix}
\mathbf{I}_{3\times 3}  &-[\textbf{T}_k\textbf{p}_k]_{\times}\\
\textbf{0}_{1\times 3} & \textbf{0}_{1\times 3}
\end{bmatrix}.
\end{aligned}
\end{equation}
Note that here $[\textbf{T}_k\textbf{p}_k]_{\times}$ transforms 4D point expression $\{x,y,z,1\}$ into 3D point expression $\{x,y,z\}$ before calculating the skew matrix. The Jacobian matrix of point-to-edge residual is defined by
\begin{equation}
\begin{aligned}
   \textbf{J}_{\mathcal{E}} &= \frac{\partial f_\mathcal{E}(\hat{\textbf{p}}_k)}{\partial \hat{\textbf{p}}_k} \cdot \textbf{J}_p\\
   &= \textbf{p}_n^T \cdot   \frac{\textbf{J}_p \times (\hat{\textbf{p}}_k-\textbf{p}^{\mathcal{E}}_1) +  (\hat{\textbf{p}}_k-\textbf{p}^{\mathcal{E}}_2) \times \textbf{J}_p }{|\textbf{p}^{\mathcal{E}}_1-\textbf{p}^{\mathcal{E}}_2|}\\
   &=\textbf{p}_n^T \cdot \frac{(\textbf{p}^{\mathcal{E}}_1-\textbf{p}^{\mathcal{E}}_2)}{|\textbf{p}^{\mathcal{E}}_1-\textbf{p}^{\mathcal{E}}_2|} \times \textbf{J}_p
   ,
\end{aligned}
\end{equation}
where 
\begin{equation}
\textbf{p}_n = \frac{(\hat{\textbf{p}}_{k} - \textbf{p}^{\mathcal{E}}_2)\times (\hat{\textbf{p}}_{k} - \textbf{p}^{\mathcal{E}}_1)}{|(\hat{\textbf{p}}_{k} - \textbf{p}^{\mathcal{E}}_2)\times (\hat{\textbf{p}}_{k} - \textbf{p}^{\mathcal{E}}_1)|}.
\end{equation}
Similarly, we can derive the Jacobian matrix of point-to-plane residual:
\begin{equation}
\begin{aligned}
   \textbf{J}_{\mathcal{S}} &= \frac{\partial f_\mathcal{S}(\hat{\textbf{p}}_k)}{\partial \hat{\textbf{p}}_k} \cdot \textbf{J}_p \\
   &= \frac{{[(\textbf{p}^{\mathcal{S}}_1 - \textbf{p}^{\mathcal{S}}_2)\times (\textbf{p}^{\mathcal{S}}_1-\textbf{p}^{\mathcal{S}}_3)]}^T}{|(\textbf{p}^{\mathcal{S}}_1 - \textbf{p}^{\mathcal{S}}_2])\times (\textbf{p}^{\mathcal{S}}_1-\textbf{p}^{\mathcal{S}}_3)|} \cdot \textbf{J}_p.
\end{aligned}
\end{equation}

\begin{algorithm}[t]
\SetAlgoLined
\SetKwInOut{Input}{Input}
\SetKwInOut{Output}{Output}
 \Input{Current scan $\mathcal{P}_{k}$, the robot trajectory $\textbf{T}_{1:k-1}$}
 \Output{Current pose $\textbf{T}_{k}$}
\Begin{
    \lIf{Not Initialized}{$\textbf{T}_{0} \leftarrow \textbf{0}$}
    Calculate the local smoothness and extract edge features and planar features \;
    Compute initial alignment $\textbf{T}_{k}^0 \leftarrow \textbf{T}_{k-1} \textbf{T}_{k-2}^{-1} \textbf{T}_{k-1}$\;
    \For{Pose optimization is not converged}{
        \For{each point $\textbf{p}_{k} \in \mathcal{P}_{k}$}{
            Transform to map coordinate 
            $\hat{\textbf{p}}_k \leftarrow \textbf{T}_{k}^{i-1}\textbf{p}_k$\;
            
            \If{$p_{k}$ is an edge feature}{
                Compute edge residual $f_{\mathcal{E}}$ and Jacobian matrix $J_{\mathcal{E}}$ \;
            }
            \If{$\mathbf{p}_{k}$ is a planar feature}{
                compute planar residual $f_{\mathcal{S}}$ and Jacobian matrix $J_{\mathcal{S}}$\;
            }
        }
        \If{nonlinear optimization converges}{
            Compute $\textbf{T}_{k}^i$\;
        }
    }    
    Update current scan into local map and remove oldest scan from local map\;
    \textbf{Return} current pose $\textbf{T}_{k}$\;
} 
\caption{Pose estimation for Solid-state LiDAR}
\label{fig: algorrithm1}
\end{algorithm}

The alignment of current scan and local map may not be ideal at the beginning. Therefore, it is necessary to search for better feature correspondences. The optimal feature correspondence can be found through an iterative manner, \textit{i.e.}, with an initial pose $\textbf{T}_k^0$ and initial correspondence based on $\textbf{T}_k^0$, we can derive the odometry estimation $\textbf{T}_k^1$ then $\textbf{T}_k^2$ and  $\textbf{T}_k^3$, \textit{etc}. This finally converges to the optimal estimation of current pose. Although iterative calculation is computationally inefficient, a good estimation of initial pose alignment is able to speed up the convergence. To find an better initial alignment, we assume constant angular velocity and linear translation: 
\begin{equation}
\begin{aligned}
  \textbf{T}_k^0 &= \textbf{T}_{k-1}\cdot \textbf{T}_{k-2}^{k-1} \\
    &=\textbf{T}_{k-1}\cdot \textbf{T}_{k-2}^{-1}\cdot \textbf{T}_{k-1}.
\end{aligned}
\end{equation}
The process of iterative odometry estimation is listed in \aref{fig: algorrithm1}.

\subsection{Probability Map Construction}
The global map is often of large size and it is not computationally efficient to update it with each frame. Therefore, we only use key frames to update and reconstruct map. The key frames are selected based on the following criterion: (1) If the displacement of robot is significant enough (\textit{i.e.}, greater than a pre-defined threshold); (2) If change of rotation angle (including roll, pitch, yaw angle change) is large; (3) If the time elapsed is more than a certain period. In practice, the rotation and translation thresholds are defined based on the FoV of the sensor, while the minimum update rate is defined based on the computational power of the processor. To increase the searching efficiency, an octree is used to construct the global map. It only takes computational complexity of $\mathcal{O}(\log  n)$ to search for a specific node from an octree of depth $n$, which can significantly reduce mapping cost \cite{hornung2013octomap}. For each cell in octree, we use $P(n|z_{1:t})$ to present the probability of the existence of an object \cite{moravec1985high}:
\begin{equation}
\begin{aligned}
  P(n|&z_{1:t}) = \\
  &{[1+\frac{1-P(n|z_{t})}{P(n|z_{t})}\cdot \frac{1-P(n|z_{1:t-1})}{P(n|z_{1:t-1})} \cdot \frac{P(n)}{1-P(n)}]}^{-1},
\end{aligned}
\end{equation}
where $z_t$ is the current measurement, $z_{1:t-1}$ is the historical measurements from key frames, $P(n)$ is the prior probability, which is set to 0.5 for unknown area. 
\begin{figure}[!t]
\begin{center}
\vspace{8pt}
\includegraphics[width=0.99\linewidth]{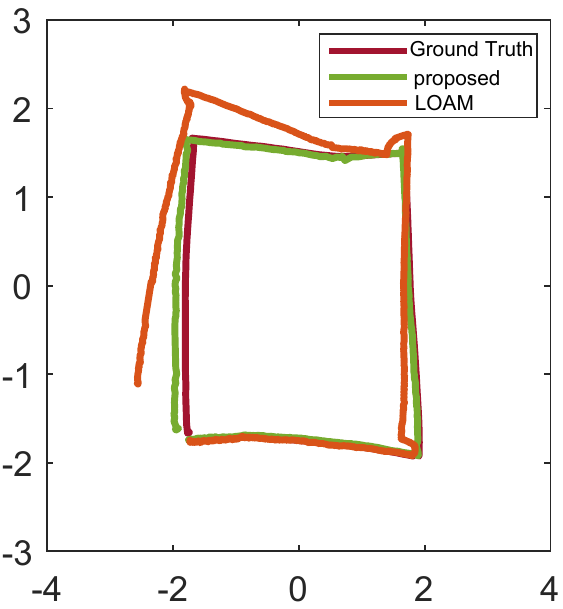}
\captionsetup{justification=justified}
\caption{Comparison among the proposed method, LOAM and ground truth. LOAM loses tracking when the rotation speed is high while the proposed method is able to track accurately. The units are in meters. }
\label{fig: vicon}
\end{center}
\end{figure}

%% file: body/Experiments.tex
\subsection{Experiment Setup}
In this section we present experimental results of the proposed method. Our method is firstly evaluated in a room equipped with a VICON system. Then, it is implemented on an Automated Guided Vehicle (AGV) that is used for warehouse manipulation. We analyze the performance of the proposed method and compare it with existing LiDAR SLAM. To further illustrate the robustness, the proposed method is also integrated into a hand-held device used for 3D scanning. In our experiment, an Intel Realsense L515 is used for demonstration. It is a small FoV solid-state LiDAR with $70^{\circ} \times 55^{\circ}$ viewing angle and 30 Hz update frequency. It is smaller and lighter than a smart phone so that it can be used in many mobile robotic platforms. The algorithm is coded in C++ and is implemented on Ubuntu 18.04 and ROS Melodic \cite{quigley2015programming}. In the first experiment, the proposed method is tested on a desktop PC with an  Intel 6-core i7-8700 processor CPU. For the experiment on the AGV and hand-held device, an Intel NUC mini computing platform is used which has an i5-10210U processor. 

\begin{figure*}[t]
\begin{center}
\vspace{8pt}
\includegraphics[width=0.99\linewidth]{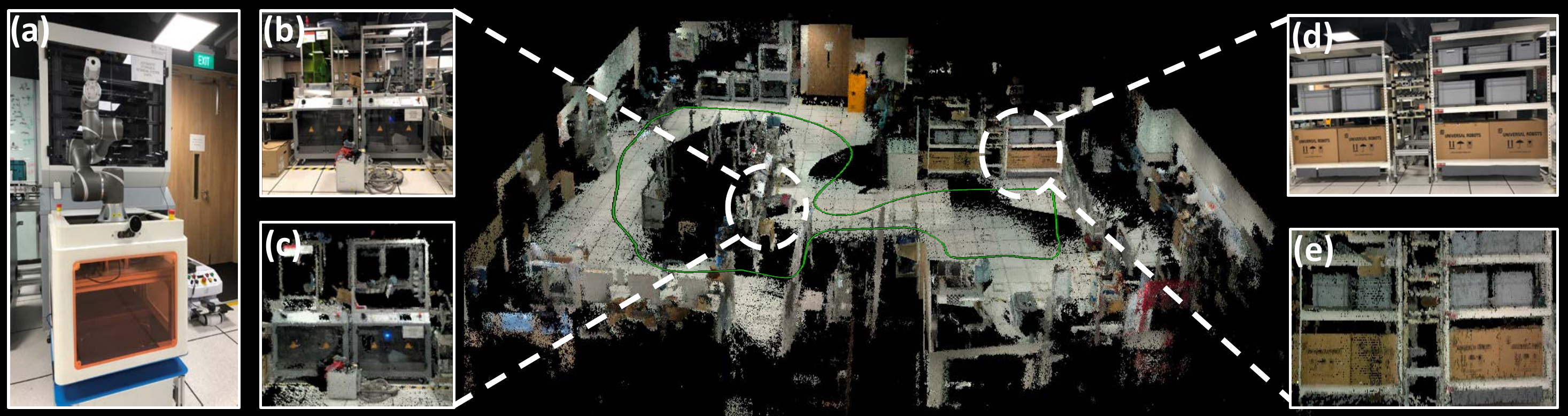}
\captionsetup{justification=justified}
\caption{Example of indoor localization and mapping in warehouse environment. (a) The AGV platform used for warehouse manipulation with a solid-state LiDAR mounted in the front. And the reconstructed map is shown in the middle. We randomly picked two places for illustration. (b) and (d) are the original camera view. (c) and (e) are reconstructed scene based on the proposed approach. The trajectory is plotted in green}
\label{fig: warehouse}
\end{center}
\end{figure*}

\subsection{Performance Evaluation and Comparison}
To evaluate the localization results, our method is compared with the ground truth, which is provided by a VICON system. A robot is manually controlled to move around the VICON room of size 4m $\times$ 4m. The result is shown in \fref{fig: vicon}, where the trajectories of ground truth and our method are plotted in red and green, respectively. The average computing time is 31 ms per frame. Our method achieves a translational error of 5\textit{cm}.
We also compared our method with LOAM \cite{zhang2014loam} that is widely used for LiDAR SLAM. The vertical angle and horizontal angle inputs in LOAM are changed according to the sensor properties of L515, while we keep the number of edge and planar features unchanged. The result of LOAM is plotted in orange. It is obvious that tracking loss happens to LOAM when there is large rotation, while our method is still able to track accurately. 

\subsection{Performance on Warehouse Robot}
The algorithm is then evaluated on an AGV running in a warehouse environment. Smart manufacturing is one of the main applications for SLAM. In an advanced factory, the robot is supposed to transport, process, and assemble products automatically. 
This requires the robot to efficiently localize itself in complex and highly dynamic environments with moving operators and other robots. In this experiment, the proposed method is integrated into an industrial AGV that is shown in \fref{fig: warehouse} (a). The robot is used for griping and transporting the materials with a robot arm on the top. A solid-state LiDAR is mounted in the front to provide localization. In this task, the AGV platform automatically explores the warehouse and reconstructs the environment at the maximum speed of 0.8 m/s. The localization and mapping results are shown in \fref{fig: warehouse} (b) with the robot trajectory plotted in green. Our method achieves an average computing time of 42 \textit{ms} on an Intel NUC mini computing platform. It can be seen that our method is able to provide reliable localization in complex environments. We compare the 3D mapping results with the images. The image view of the original warehouse is shown in \fref{fig: warehouse} (b), (d) and the built 3D map is shown in \fref{fig: warehouse} (c), (e). To evaluate the quality of mapping, we compare the built objects with actual objects. Specifically, we measure the size of operating machines which are shown in Fig.4 (b) and Fig.4 (d). The edge points from built object are picked and the Euclidean distances between edge points are measured. The results are calculated by taking the average of the measured distances. The measured dimension and actual size of the operation machine in \fref{fig: warehouse} (b) are $1.17m\times 1.88m $ and $1.15m\times 1.85m$, while the measured dimension and actual size of operation machine in \fref{fig: warehouse} (d) are $1.16m\times 1.94m$ and $1.16m\times 1.95m$ . It can be seen that our approach is also able to build a high resolution 3D map that is close to real environment.

\begin{figure}[t]
\begin{center}
\vspace{15pt}
\includegraphics[width=0.99\linewidth]{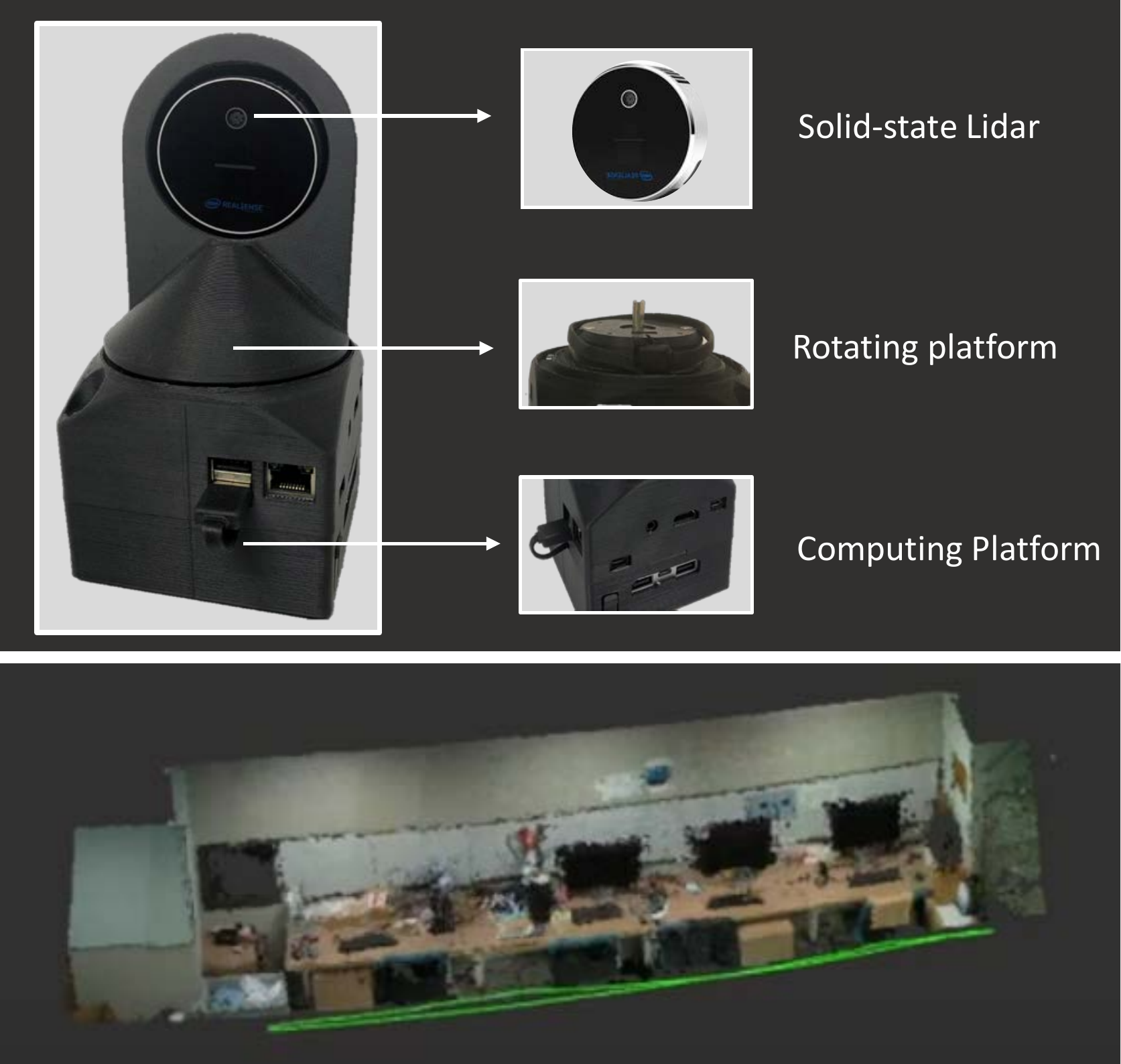}
\captionsetup{justification=justified}
\caption{Integration of the proposed method on a hand-held scanner. (a) Light weight hand-held scanner using solid-state LiDAR as perception system. (b) Localization and mapping result, the trajectory is plotted in green.}
\label{fig: scan_graph}
\end{center}
\end{figure}

\subsection{Performance on Hand-held Device}
\subsubsection{3D Mapping} To further demonstrate the robustness, the proposed method is also evaluated on a hand-held device. With the growing of Virtual Reality (VR), Augmented Reality (AR), and gaming industry, SLAM has been implemented on various mobile devices such as smart phones and VR glasses. However, most mobile platforms have limited computational resources. Compared to implementation on warehouse robot, the hand-held device suffers from vibration and large viewing angle change which can cause tracking loss and localization failure. Our system is built as a mobile scanner which is shown in \fref{fig: scan_graph}(a). The hand-held scanner consists of three modules: perception module, rotation platform, and computing module. It is less than 500 grams and can be implemented on many applications such as drone navigation. In the experiment, we hold the 3D scanner and scan the indoor environment at normal walking speed. The localization and mapping result is shown in \fref{fig: scan_graph}(b), with the trajectory plotted in green. Our method can accurately localize itself and perform mapping in real time. 

\subsubsection{Rotation Test} The hand-held device often has higher rotation changes which can result in tracking loss. In order to demonstrate the performance of the proposed method under large rotation, we put the solid-state LiDAR in horizontal orientation and randomly rotate the solid-state LiDAR with maximum rotation speed of 1.57 rad/s. The solid-state LiDAR returns to horizontal orientation and we record the angle deviation. Tracking loss is considered when the final angle deviation is greater than 10 degrees. We compare our method with A-LOAM and the results are shown in \tref{table:rotation test}. It can be seen that our method has a higher success rate compared to A-LOAM. 

\begin{table}[t]
    \begin{center}
    \begin{tabular}{ccc}
    \toprule
    Method & Success & Tracking loss  \\
    \midrule
    The proposed method  &6 & 0   \\ 
    A-LOAM  &1 & 5    \\ 
    \bottomrule
    \end{tabular}
    \caption{Results of rotation test. }
    \label{table:rotation test}
    \end{center}
\end{table}

%% file: body/Conclusion.tex
In this paper, we present a full SLAM framework for solid-state LiDAR which is an emerging LiDAR system with higher update frequency and smaller FoV compared to traditional mechanical LiDAR. Our system mainly consists of rotation invariant feature extraction, odometry estimation, and probability map construction. The propose method is able to support real time localization and densed mapping on an embedded mini PC. Thorough experiments have been performed to evaluate the proposed method, including experiments on a warehouse AGV and a hand-held mobile device. The results show that the proposed method is able to provide reliable and accurate localization and mapping at high frequency. It can be implemented on most of mobile platform such as UAV and hand-held scanner. We have made the source codes publicly available.